\documentclass[10pt,twocolumn,letterpaper]{article}

\usepackage{cvpr}
\usepackage{times}
\usepackage{epsfig}
\usepackage{graphicx}
\usepackage{amsmath}
\usepackage{amssymb}
\usepackage{multirow}
\usepackage{booktabs}

\usepackage[dvipsnames]{xcolor}
\usepackage{float}

\newcommand{\parahead}[1]{\par\textbf{#1}:\ }

\newcommand{\filluptopage}[1]{%
  \clearpage
  \loop\ifnum\value{page}<#1\relax
    \null\clearpage
  \repeat
  \loop\ifnum\value{page}=#1\relax
    \null\clearpage
  \repeat
}

\usepackage[pagebackref=true,breaklinks=true,letterpaper=true,colorlinks,bookmarks=false]{hyperref}

\cvprfinalcopy 

\ifcvprfinal\fi
\begin{document}

\title{Learning Generalizable Physical Dynamics of 3D Rigid Objects}

\author{Davis Rempe \qquad Srinath Sridhar \qquad He Wang \qquad Leonidas J. Guibas\\Stanford University
}

\maketitle
\thispagestyle{empty}

\begin{abstract}
  Humans have a remarkable ability to predict the effect of physical interactions on the dynamics of objects. Endowing machines with this ability would allow important applications in areas like robotics and autonomous vehicles. In this work, we focus on predicting the dynamics of 3D rigid objects, in particular an object's final resting position and total rotation when subjected to an impulsive force. Different from previous work, our approach is capable of generalizing to unseen object shapes---an important requirement for real-world applications. To achieve this, we represent object shape as a 3D point cloud that is used as input to a neural network, making our approach agnostic to appearance variation. The design of our network is informed by an understanding of physical laws. We train our model with data from a physics engine that simulates the dynamics of a large number of shapes. Experiments show that we can accurately predict the resting position and total rotation for unseen object geometries. 
\end{abstract}
\section{Introduction}
\label{sec:intro}
Humans have a fundamental intuitive understanding of the dynamics of the physical world. Even at a young age, we are able to understand and predict the effect of physical interactions with objects~\cite{baillargeon1990top,leslie1982perception} (\eg, putting a peg into a hole, catching a ball). This \emph{intuitive knowledge of dynamics} allows us to operate in previously unseen environments, and interact with and manipulate objects encountered for the first time. Endowing machines with the same ability would allow new applications in autonomous driving, home robotics, and augmented reality (AR) scenarios. 

The 3D dynamics of objects can be predicted using well-studied physical laws given precise properties and system parameters (\eg, mass, moment of inertia, applied force). In practice however, it is impossible to estimate all system parameters, especially from non-contact sensory data. Furthermore, simulating the physics of complex environments requires exact specification of a partially-observed system, and can be computationally expensive and imprecise.

Inspired by the generalizable ability of humans to intuit object dynamics, we develop a deep learning approach to predict the physical dynamics of unseen 3D rigid objects. Learned dynamics has advantages over traditional simulation as it offers differentiable predictions useful for reinforcement learning, and the flexibility to tradeoff speed and accuracy by varying network capacity. There has recently been a lot of interest in learning to predict object dynamics, yet, a number of limitations remain. First, prior work lacks the ability to generalize to shapes that were not seen at training time~\cite{byravan2017se3}, limiting real-world applicability.
Second, many methods are limited to 2D objects and environments ~\cite{battaglia2016interactionnets,chang2017compositional,fragkiadaki2016visualbilliards,watters2017vin} and do not generalize well to 3D objects. 
Finally, many methods use images as input~\cite{mottaghi2016if,mottaghi2016newton,finn2016videoprediction,agrawal2016poke} which provide only partial shape information and entangle variations in object appearance with physical motion.

\begin{figure*}[!ht]
\begin{center}
   \includegraphics[width=\linewidth]{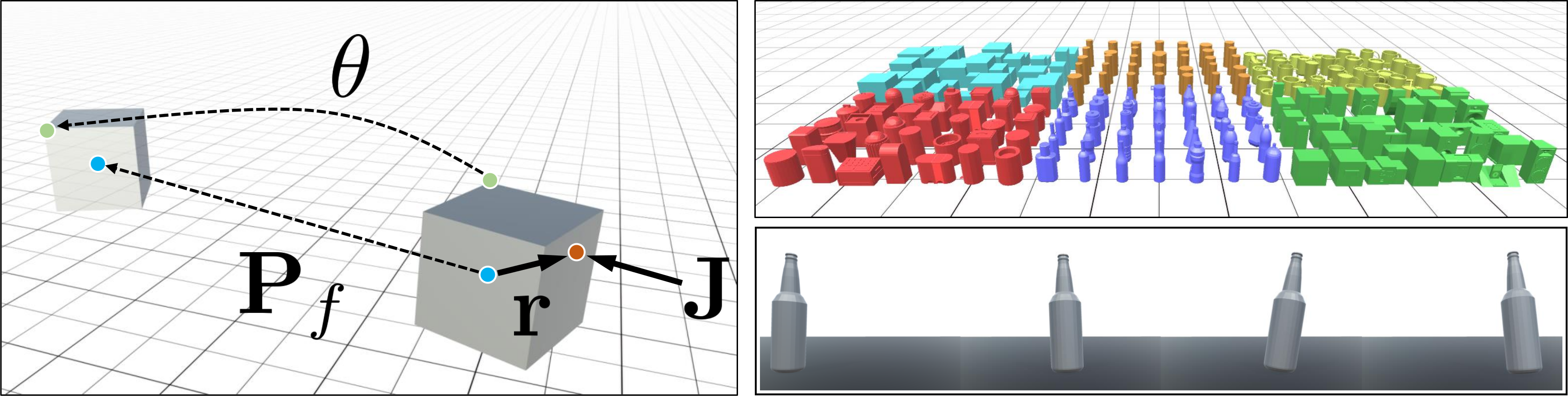}
\end{center}
   \caption{We study the problem of predicting the \emph{position} $\mathbf{P_f}$ and total \emph{rotation} $\theta$ of an object initially resting on a plane subjected to an impulse $\mathbf{J}$ at position $\mathbf{r}$ (left). Our method can predict the dynamics of a variety of different shapes (right top) and generalizes to previously unseen object shapes and impulses. Our problem formulation presents many challenges including the unpredictable 3D motion caused due to \emph{wobbling} of objects under motion (right bottom). We are able to accurately predict dynamics even under these conditions.
   }
\label{fig:problem}
\end{figure*}

Our goal is to learn to predict the dynamics of 3D rigid objects and generalize these predictions to previously unseen object geometries. To this end, we focus on the problem of accurately predicting the final rest state (position and total rotation) of an object (initially stationary on a plane) that has been subjected to an \emph{impulse}---a force causing an instantaneous change in velocity. As a result of this impulse, the object moves along the plane but friction eventually brings it to rest (see Figure~\ref{fig:problem}). This problem formulation has surprisingly many nuances. The motion of an object after an applied impulse depends non-linearly on factors such as its moment of inertia, amplitude of the force, and surface friction. Furthermore, sliding objects could wobble resulting in unpredictable motions. Learning these subtleties in a generalizable way requires a deep understanding of the connection between object shape, mass, and dynamics. Since this problem formulation is well-defined, it allows us to better evaluate shape generalization without worrying about other complex dynamics like collisions. Yet, it still has practical applications, for instance, in robotic pushing of objects, and is a strong foundation for developing methods to predict more complex physical dynamics.

To solve this problem, we present a neural network model that takes the shape of an object and additional information about the applied impulse as the input, and predicts the final rest position and \emph{total rotation} undergone throughout the entire motion of the object. Different to previous work, we use a 3D point cloud to represent the shape of the object and use features extracted by PointNet~\cite{qi2017pointnet}. This makes our method more robust and applicable to the real world since we decouple object motion from appearance variation. Furthermore, our network design is informed by an understanding of real-world physical laws and priors. To train this network, we simulate the physics of a large number of household object shapes from the ShapeNet repository~\cite{chang2015shapenet}. Our network learns to extract salient shape features from these examples. This allows it to make accurate predictions not just for impulses and object shapes seen during training, but also for unseen objects in novel shape categories that are subjected to new impulses.

We present extensive experiments that demonstrate that our method is capable of learning physical dynamics that generalize to unseen 3D object shapes. We compare the performance of our approach with a standard physics engine as well as other state-of-the-art approaches~\cite{mrowca2018flexible}. We offer additional insights into the meaningful features learned by our model and justify our design decisions.
\section{Related Work}
\label{sec:relwork}
The problem of \emph{learned physical understanding} has been approached in many ways, resulting in multiple formulations and ideas of what it means to \emph{understand} physics. Some work answers questions related to physical aspects of a scene~\cite{battaglia2013simunderstanding,zhang2016blocks,li2016fall,li2017stability,lerer2016fbtowers,mirza2016unsupervised}, while others learn to infer physical properties of objects from video frames~\cite{wu2015galileo,wu2016phys101,wu2017deanimation} or image and 3D object information~\cite{liu2018ppd}. In this section we focus on describing work most closely related to ours.

\parahead{Forward Dynamics Prediction}
Many methods attempt direct forward prediction of object dynamics. Generally, these approaches take the current state of objects in a scene, the state of the environment, and any external forces as input and predict the state of objects at future times. Forward prediction is a desirable approach as it can be used for action planning~\cite{hamrick2016decision}. Multiple methods have shown success in 2D settings~\cite{fraccaro2017disentangled}. \cite{fragkiadaki2016visualbilliards} uses raw visual input centered around a ball on a table to predict multiple future positions. The \emph{neural physics engine}~\cite{chang2017compositional} and \emph{interaction network}~\cite{battaglia2016interactionnets} explicitly model relationships in a scene to accurately predict the outcome of complex interactions like collisions between balls. \cite{watters2017vin} builds on~\cite{battaglia2016interactionnets} by adding a front-end perception module to learn a state representation for objects. These 2D methods exhibit believable results, but are limited to simple primitive objects. They also make predictions for a small number of future frames rather than final rest state. As a result, they must iteratively roll out predictions to make observations far into the future, providing detailed object trajectories at the cost of compounding error. 

\parahead{Dynamics in Images \& Videos}
Many methods for 3D dynamics prediction operate on RGB images or video frames~\cite{ye2018interpretable,riochet2018intphys,ehrhardt2017mechanics,ehrhardt2017longterm,stewart2017label,ehrhardt2018visualobs}. \cite{mottaghi2016newton} and \cite{mottaghi2016if} introduce multiple algorithms to infer future 3D translations and velocities of objects given a single static RGB image. Some methods directly predict pixels of future frames conditioned on actions~\cite{oh2015atari}. \cite{finn2016videoprediction} infers future video frames involving robotic pushing conditioned on the parameters of the push and uses this prediction to plan actions~\cite{finn2017planning}. In a similar vein, \cite{agrawal2016poke} uses video of a robot poking objects to implicitly predict object motion and perform action planning with the same robotic arm. Many of these methods focus on real-world settings, but do not use 3D information and possibly entangle object appearance with physical properties.

\parahead{3D Physical Dynamics}
Recent work has taken initial steps towards more general 3D settings~\cite{wang2018physnet}. Our method is most similar to \cite{byravan2017se3} who use a series of depth images to identify rigid objects and predict point-wise transformations one step into the future, conditioned on an action. However, they do not show generalization to unseen objects. Other work extends ideas introduced in 2D by using variations of graph networks. \cite{sanchez2018graphnet} decomposes systems containing connected rigid parts into a graph network of bodies and joints to make single-timestep forward predictions. The hierarchical relation network (HRN)~\cite{mrowca2018flexible} breaks rigid and deformable objects into a hierarchical graph of particles to learn particle relationships and dynamics from example simulations. Forward predictions for each of these particles result in motion of objects in the scene. Though HRN is robust to novel objects, it is unclear whether it can generalize to real-world scenarios due to detailed per-particle supervision required during training (see Section~\ref{sec:results}).
\section{Problem Formulation}
We investigate the problem of predicting the dynamics of an initially stationary rigid object subjected to an impulse. We assume the following \textbf{inputs}: (1)~the shape of the object in the form of a point cloud ($\mathbf{O} \in \mathbb{R}^{N\times 3}$), and (2)~the applied impulse vector and its position. We further assume that the object moves on a plane under standard gravity, the applied impulse is parallel to the plane at the same height as the center of mass (see Figure~\ref{fig:formulation_detail}), and the friction coefficient between the plane and the object is constant.

Our goal is to accurately \textbf{predict} the final rest position ($\mathbf{P}_f \in \mathbb{R}^2$) and the total rotation ($\theta \in \mathbb{R}$) (about the vertical axis) of an object subjected to an impulse. Since the object could undergo multiple 360$^\circ$ rotations before coming to rest, the total rotation $\theta$ is often different from the rigid rotation. While we parameterize the final object state with 2 translational and 1 rotational parameters, we \emph{do not} restrict the object motion to 2D. As shown in Figure~\ref{fig:problem}, the object is free to move in 3D as long as it does not topple over. We use a point cloud to encode object geometry since it only depends on the surface geometry, making it agnostic to appearance, and can be readily captured in a real-world setting through commodity depth sensors. Additionally, the applied impulse vector and position could be acquired through the agent executing the action (\eg, a robotic arm).

Instead of solving the highly challenging unconstrained 3D dynamics prediction problem, we choose to specifically model 3D motion along a plane and predict final rest state (as opposed to multi-step~\cite{mrowca2018flexible}). Since we predict final state rather than a sequence, we apply an instantaneous impulse rather than a time-varying force. We do not allow the object to roll on its side or to topple over, but 3D \emph{wobbling motion} (see Figure~\ref{fig:problem}) causes the amount of contact surface area to vary resulting in complex trajectories. Unobserved quantities (\eg, mass, volume, moment of inertia, and contact surface) additionally contribute to the difficulty of this problem. Learning the dynamics of 3D planar motion has many benefits. It allows us to work on a well-defined problem and to focus on gaining insight and evaluating generalization to unseen object shapes without complex interactions such as collisions. Furthermore, such a formulation has practical applications, for instance a robotic arm pushing objects on a desk to reach a goal state. A solution to this 3D problem would also be a strong foundation to predicting more complex dynamics.

What does a network need to learn to accurately predict final rest state? Assuming constant friction, the final rest state of an object depends on its initial linear and angular velocities defined as
\begin{align}
\mathbf{v} &= \frac{\mathbf{J}}{m}, &
\omega &= \frac{\mathbf{r} \times \mathbf{J}}{I},
\label{eqn:laws}
\end{align}
where $\mathbf{v}$ is the linear velocity, $\mathbf{J}$ is the applied impulse vector, $m$ is the object mass, $\omega$ is the angular velocity, $\mathbf{r}$ is the impulse position relative to the center of mass, and $I$ is the moment of inertia. The final position of the object is proportional to the square of the starting linear velocity---a straightforward relationship to capture with neural networks. However, the final rotation, even in our setup, is non-linear and depends on factors such as object mass distribution, friction, and the contact surface area and shape. Furthermore, out-of-plane motions lead to uneven friction causing non-linear translation.

\section{Data Simulation}
\label{sec:data}
We use 3D simulation data from the Bullet physics engine~\cite{bullet} within Unity~\cite{unity} for our problem, however, our method in principle could be trained on real-world data provided ground truth shape (\eg, from a depth sensor), impulse vector (\eg, from a robotic arm), and translation and total rotation (\eg, from visual tracking) is available.

A single datapoint in our datasets is a unique simulation run in which an object is placed at rest on a flat plane and then a random impulsive force is applied to the surface of the object in a direction parallel to the ground plane. As a result, the object moves and eventually comes to rest at its final position and total rotation. For each simulation, we record the point cloud shape of the object, the magnitude, direction and position of the applied impulse, and the final object resting position and total rotation.
\begin{figure}[!ht]
\begin{center}
   \includegraphics[width=\linewidth]{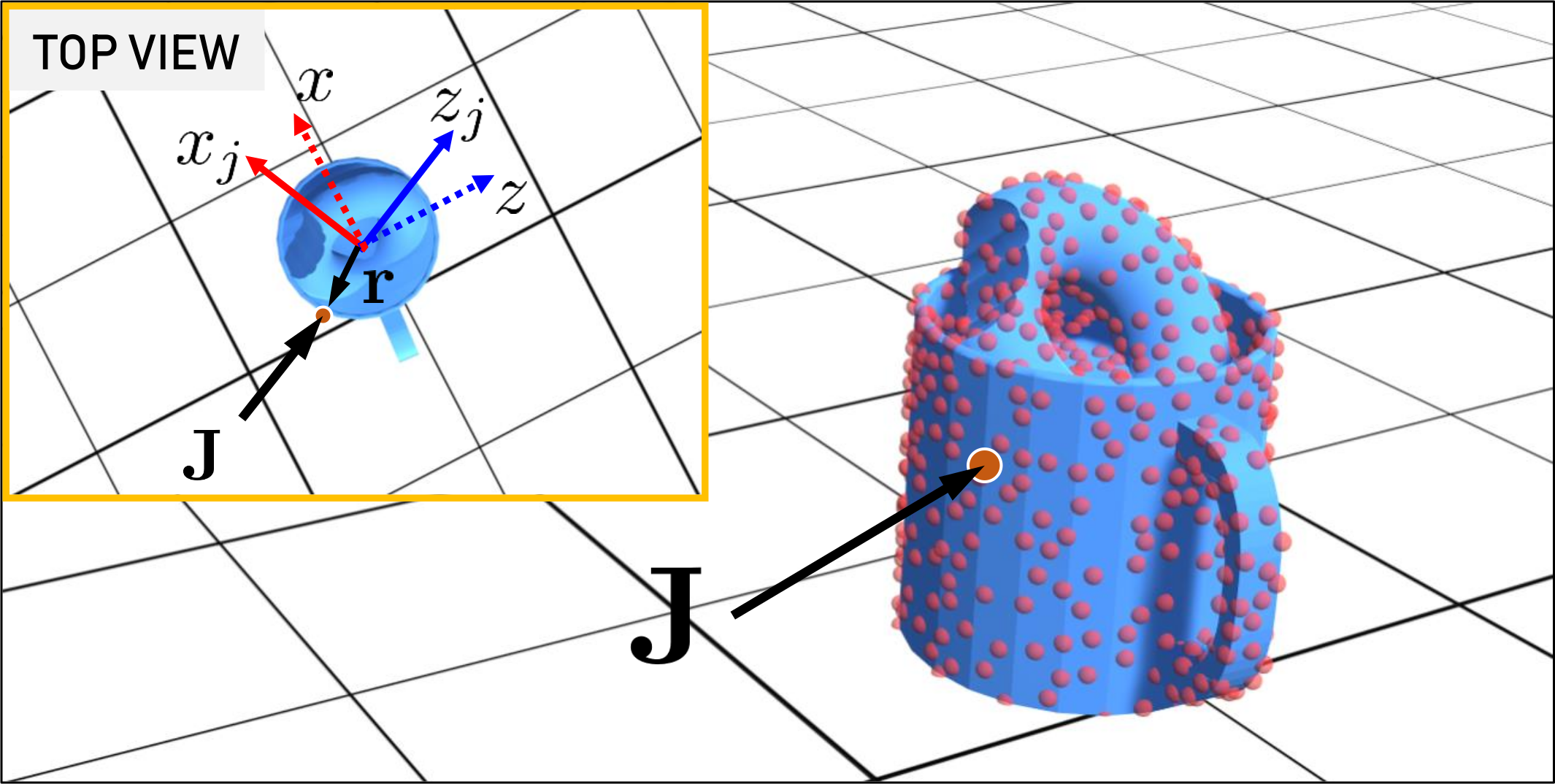}
\end{center}
   \caption{Problem input and impulse coordinates. Our method uses point clouds (red spheres) of object shapes, the impulse vector $\mathbf{J}$ (black arrow), and the impulse position (red circle) to learn and predict dynamics. The object is initially at rest on a plane. 
   We use \emph{impulse coordinates}, aligned with the direction of the impulse, to represent the input to our model.
   }
\label{fig:formulation_detail}
\end{figure}

\parahead{Simulation Procedure}
For our input, we sample a point cloud with 1024 points from the surface of each unique object in all datasets (see Figure~\ref{fig:formulation_detail}).
The applied impulse direction, magnitude, and position are chosen randomly from a uniform distribution. We hold the ground and object friction values and object density constant across all simulations. During simulation, we use the exact mesh to build a collider that captures the object geometry complexity to simulate motion and contact with the ground plane. We ensure that simulated objects do not fall over, however motion is not explicitly constrained in any way. Simulated objects range from 0.15 to 20~kg in mass, typically travel between 0.5 and 5 meters, and can rotate from 0$^\circ$ to more than 2000$^\circ$ (5--6 complete rotations).

\begin{figure}[t]
\begin{center}
   \includegraphics[width=\linewidth]{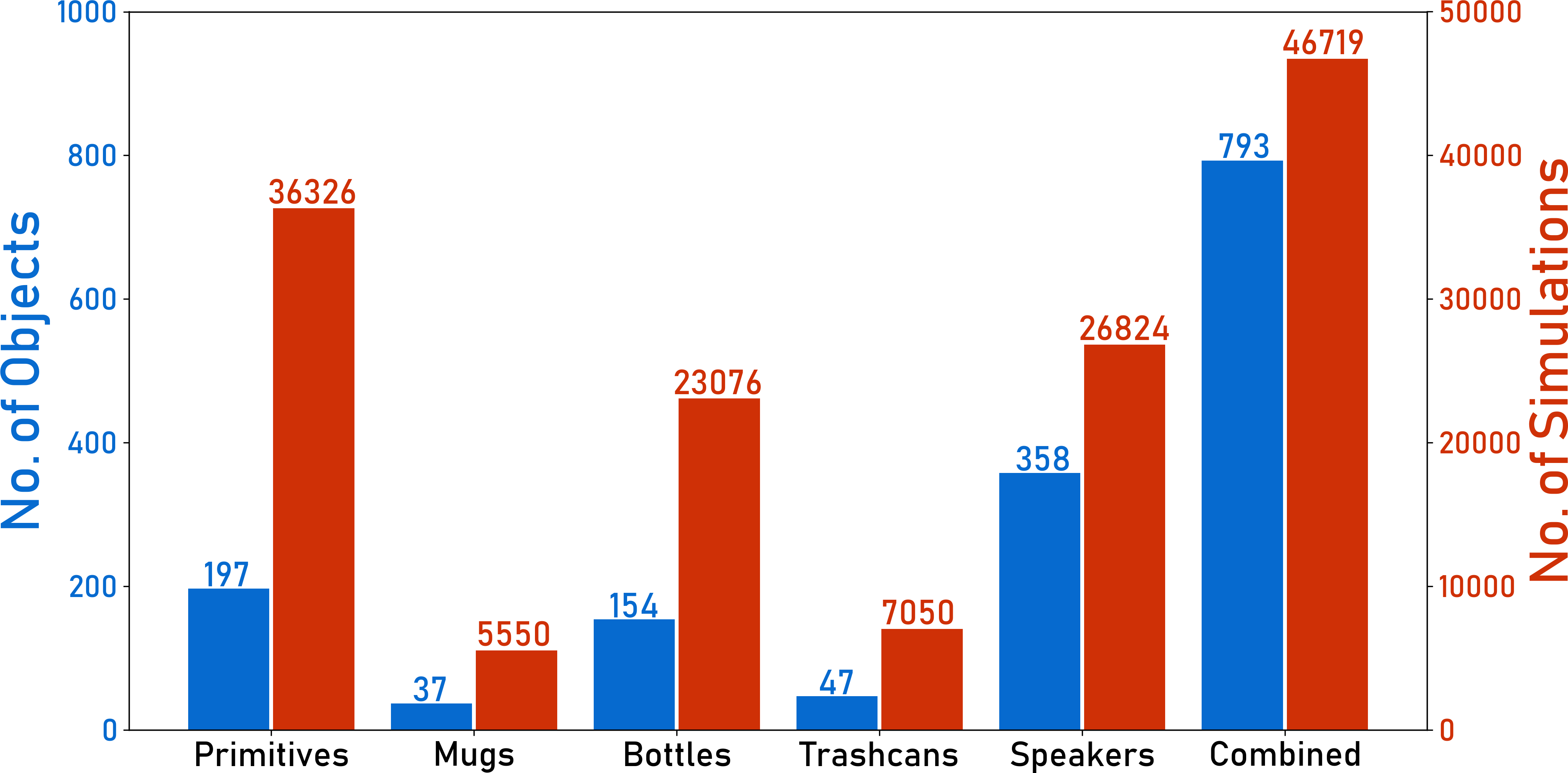}
\end{center}
   \caption{Data distribution. We generated our data using 6 categories of shapes both from ShapeNet and primitive shapes (\texttt{Box} and \texttt{Cylinders} are combined into \texttt{Primitives}). For each shape category, we ran thousands of simulations. In total, we use 793 unique object shapes and ran 98826 simulations.}
\label{fig:datadist}
\end{figure}

\parahead{Datasets}
We synthesized multiple categories of datasets to train and evaluate our models. These separate datasets contain different sets of objects and are summarized in Figure~\ref{fig:datadist}. Each dataset is split into training (80\%) and test sets (20\%). Training objects are simulated with a different random scale from 0.5 to 1.5 for x, y, and z directions in order to increase shape diversity. There are two primitive object datasets used for evaluating on relatively low shape diversity. The \texttt{Box} dataset is a single cube, whereas the \texttt{Cylinders} set contains a variety of cylinder shapes. There are four datasets which contain everyday objects taken from the ShapeNet~\cite{chang2015shapenet} dataset. These exhibit wide shape diversity and offer a more challenging task. Lastly, we have a dataset which combines all of the objects from the previous six to create a large and extremely diverse set of shapes. This \texttt{Combined} dataset is split roughly evenly between shape categories (around 7000--10000 simulations per category). In total, we use \textbf{793} distinct object shapes and ran \textbf{98826} simulations to generate our data.

\parahead{Pre-processing}
Outlier simulations where the object translated more than 7 meters or rotated more than 3000$^\circ$ are removed. We also transform all data to a coordinate system where the x-axis is aligned with the direction of the applied impulse---we refer to these as \textbf{impulse coordinates} (see Figure~\ref{fig:formulation_detail}). Motion will be most prominent along the direction of the impulse, thus using impulse coordinates allows our network to better learn object dynamics.

\begin{figure*}[!th]
\begin{center}
   \includegraphics[width=\textwidth]{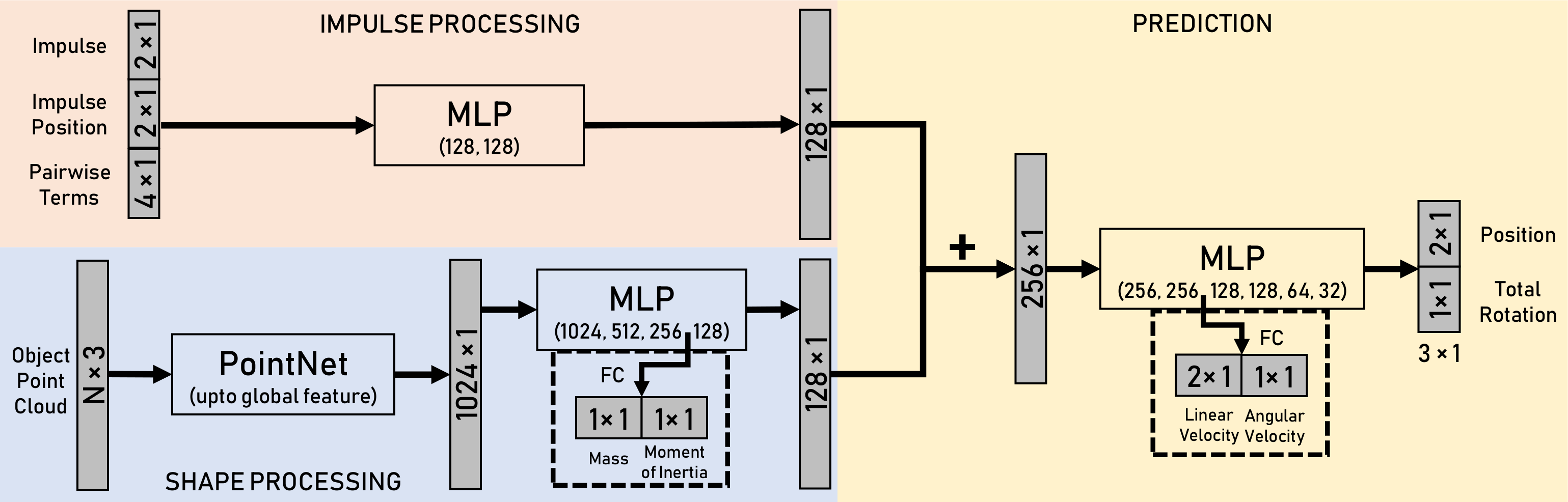}
\end{center}
   \caption{Model architecture. Our network takes the impulse, its position, additional pairwise terms, and the object point cloud as input and predicts the final resting position and total rotation that the object undergoes. Since the object's initial and final position are on the plane, we use only 2 translation and 1 rotation parameters. Numbers in bracket indicate the output size of each layer, + sign is concatenation, MLP indicates multilayer perceptron, and FC indicates a fully-connected layer. Optional branches are shown in dashed boxes.
}
\label{fig:arch}
\end{figure*}
\section{Method}
\label{sec:method}

To predict final rest position and total rotation after an impulse, we use a neural network trained on simulated data. 

\subsection{Network Architecture}
A straightforward approach may combine all inputs into one vector and use a multi-layer neural network to directly regress the final state. As we show in Section~\ref{sec:results}, this na\"ive approach cannot learn the intricacies of the non-linear motion of the object before it comes to rest. We take a more principled approach and inform the design of our network based on our understanding of physical laws and priors.

From Equation~\ref{eqn:laws}, we observe that the linear and angular velocities depend on: (1)~the applied impulse magnitude, direction, position, and its angular impulse ($\mathbf{r} \times \mathbf{J}$), and (2)~the shape of the object which affects its mass $m$ and moment of inertia $I$. We therefore base our network design on learning important information related to the applied impulse and shape of the object. Our network (see Figure~\ref{fig:arch}) is composed of two main branches whose output features are jointly used to make a final position and rotation prediction.

\parahead{Impulse Processing}
The top branch in our network is the impulse processing branch which takes the applied impulse, its position, and 4 pairwise terms as input, and outputs an \emph{impulse feature}. The 4 pairwise terms are the products of the components of the impulse with those of the impulse position $\mathbf{r}$. The aim of this branch is to learn the effect of the impulse and the angular impulse on the motion of the object. Since the impulse is parallel to the ground, we only provide the 2D impulse vector and its position relative to the center of mass. We observe that the angular impulse is a cross product ($\mathbf{r} \times \mathbf{J}$) which could be difficult to learn. We encourage the network to learn this relationship by providing the pairwise product between $\mathbf{r}$ and $\mathbf{J}$.

\parahead{Shape Processing}
The bottom shape processing branch is designed to extract salient shape features that are crucial to making accurate predictions. As seen in Equation~\ref{eqn:laws}, object geometry affects both linear and angular velocities through its mass (which depends on volume) and moment of inertia about the vertical axis. The aim of this branch is to help the network develop notions of volume, mass, and inertia from a point cloud representation. It must also learn the effect of the area and shape of the bottom contacting surface which determines how friction affects total rotation. To effectively learn this, we use PointNet~\cite{qi2017pointnet}. As shown in Figure~\ref{fig:arch}, the initial object point cloud is fed to the PointNet classification network which outputs a global feature that is further processed to output our final \emph{shape feature}. We use batch normalization following every layer in the network besides the final output layer.

\parahead{Prediction}
After concatenating the impulse and shape features, we jointly predict final position and total rotation with a 6-layer multilayer perceptron (MLP). Position is a 2$\times$1 vector and rotation a 1$\times$1 vector since the object rests on the plane in its final state. Because rotation affects final position, jointly predicting them with the same network provides improved performance.

\parahead{Optional Branches}
We add two optional branches (dashed boxes in Figure~\ref{fig:arch}) that are only used to investigate if our network learns notions of mass, moment of inertia, and linear and angular velocities (see Section~\ref{sec:results}). Each optional branch is a single fully-connected layer. The first branch takes a feature from the shape processing branch to predict the mass and moment of inertia from the point cloud. The second predicts initial linear and angular velocities from a feature in the final prediction MLP.

\subsection{Loss Functions \& Training}
The goal of the network is to minimize the error between the predicted and ground truth position and rotation. We found that using an $L^2$ loss for translation and rotation caused the network to focus too much on examples with large error due to their large translation and total rotation. Instead we propose to use a form of relative error: for translation we penalize the relative distance between the predicted final position $\mathbf{\hat{P}_f}$ and ground truth $\mathbf{P_f}$, and for rotation we use a relative $L^1$ error between the predicted total rotation $\hat{\theta}$ and the ground truth $\theta$. We sum the values in the denominator of the rotation loss $\mathcal{L}_\theta$ to avoid exploding losses when ground truth rotation is near zero:
\begin{align}
    \mathcal{L}_p &= \frac{\| \mathbf{P_f} - \mathbf{\hat{P}_f}\|}{\| \mathbf{P_f} \|}, & \mathcal{L}_\theta &= \frac{|\hat{\theta} - \theta|}{|\hat{\theta}| + |\theta|}.
    \label{eqn:loss}
\end{align}
Our final loss is the sum of the two $\mathcal{L} =  \mathcal{L}_p + \mathcal{L}_\theta$.

We train all branches of our network jointly using the Adam~\cite{kingma2015adam} optimization algorithm with a starting learning rate of 0.005 which is exponentially decayed to \mbox{1 $\times$ 10$^{-5}$} during training. In the shape processing branch, PointNet weights are pretrained on ModelNet40~\cite{wu2015modelnet}, then fine-tuned during our training process. We train the network for 200 epochs with a batch size of 128 on a single NVIDIA Titan X GPU. Before training, 20\% of the objects in the training split are set aside as validation data. During training, evaluation is performed on the validation set every 5 epochs. The model weights which result in the lowest validation loss throughout training are used as the final model. In total, our network architecture has about 2.8 million parameters.
\begin{table*}[!ht]
\begin{center}
\scalebox{0.77}{
\begin{tabular}{l l l l l l l l l l l}
\toprule
 & \multicolumn{2}{l}{\textbf{ImpulseGen, Single}} & \multicolumn{2}{l}{\textbf{ImpulseGen, Combined}} & \multicolumn{2}{l}{\textbf{ObjGen, \textcolor{blue}{Single}}} & \multicolumn{2}{l}{\textbf{ObjGen, \textcolor{orange}{Combined}}} & \multicolumn{2}{l}{\textbf{ObjGen, \textcolor{ForestGreen}{Leave-One-Out}}} \\
\midrule
Dataset & Position & Rotation & Position & Rotation & Position & Rotation & Position & Rotation & Position & Rotation \\
\midrule 
Box & 2.8 {\small(4.5)}  & 10.9 {\small(8.7)} & 4.5 {\small(8.1)} & 9.3 {\small(7.1)} & 2.8 {\small(4.5)} & 10.9 {\small(8.7)} & 3.3 {\small(5.0)} & 9.3 {\small(7.2)} & \textbf{4.1} {\small(6.6)} & \textbf{11.1} {\small(7.9)} \\ 
Cylinders & 5.4 {\small(10.0)}  & 12.2 {\small(50.5)} & 6.4 {\small(12.0)} & 16.2 {\small(56.8)} & 5.9 {\small(11.0)} & 11.5 {\small(48.5)} & 6.4 {\small(11.9)} & 17.7 {\small(75.7)} & \textbf{8.5} {\small(14.9)} & \textbf{17.7} {\small(72.1)} \\
Mugs & 3.6 {\small(6.4)} & 7.2 {\small(11.2)} & 4.2 {\small(7.1)} & 8.9 {\small(14.8)} & 11.8 {\small(20.2)} & 10.7 {\small(16.9)} & 8.1 {\small(14.1)} & 10.8 {\small(15.6)} & \textbf{8.9} {\small(15.8)} & \textbf{13.5} {\small(21.9)} \\
Trashcans & 3.9  {\small(6.7)} & 9.2 {\small(12.6)} & 4.7 {\small(8.5)} & 9.6 {\small(13.7)} & 6.4 {\small(11.4)} & 11.0 {\small(15.1)} & 6.2 {\small(11.1)} & 13.0 {\small(19.4)} & \textbf{5.6} {\small(10.0)} & \textbf{12.9} {\small(16.0)} \\
Bottles & 3.3 {\small(6.2)} & 5.6 {\small(21.5)} & 6.2 {\small(11.1)} & 12.7 {\small(42.6)} & 10.2 {\small(19.1)} & 14.7 {\small(72.0)} & 9.3 {\small(16.9)} & 15.6 {\small(62.1)} & \textbf{17.4} {\small(28.3)} & \textbf{24.0} {\small(81.7)}  \\
Speakers & 10.0 {\small(21.6)} & 14.6 {\small(63.5)} & 9.1 {\small(20.0)} & 11.4 {\small(47.5)} & 11.3 {\small(24.9)} & 13.3 {\small(51.9)} & 8.9 {\small(19.4)} & 11.4 {\small(43.4)} & 42.7 {\small(86.1)} & 160.5 {\small(586.1)}  \\
Combined & - & - & \textbf{6.2} {\small(11.9)} & \textbf{11.8} {\small(34.4)} & - & - & \textbf{6.9} {\small(13.0)} & \textbf{13.2} {\small(39.5)} & - & - \\
\bottomrule
\end{tabular}}
\end{center}
\caption{Results for impulse generalization (ImpulseGen) and object generalization (ObjGen) experiments. For position, \textbf{mean relative \% error} (and \textbf{absolute centimeter error} in parentheses) is reported. For rotation, \textbf{mean relative \% error} (and \textbf{absolute degree error)} is shown. \emph{Single} indicates a different model trained for each shape category, \emph{Combined} indicates a single model trained on the \texttt{Combined} dataset, and \emph{Leave-One-Out} indicates models trained on the \emph{Combined} dataset with the evaluated category left out.} 
\label{table:resultssummary}
\end{table*}
\section{Experiments}
\label{sec:results}
In this section, we present extensive experiments and evaluation on the generalization capability of our approach, justify design choices through ablation studies, and present comparison to baselines as well as previous work.

\parahead{Evaluation Metrics}
For all experiments, we report mean relative and absolute errors for position and total rotation. Since absolute errors increase for large motions and datasets have different distributions, relative error is the better indicator of model performance making different datasets comparable. For position, we use the same relative error used for the loss (Equation~\ref{eqn:loss}). For rotation, we report a \emph{binned} relative error
\begin{align}
    \eta_\theta = \frac{|\hat{\theta} - \theta| / b}{\lceil |\theta| / b\rceil}.
\end{align}
$\theta$ is the ground truth total rotation and $\hat{\theta}$ is the prediction. For all results we use a bin of \mbox{$b = 30^\circ$}. This metric prevents relative rotation error from exploding when ground truth rotation is near zero but still penalizes poor predictions.

\begin{figure*}[t]
\begin{center}
   \includegraphics[width=0.98\textwidth]{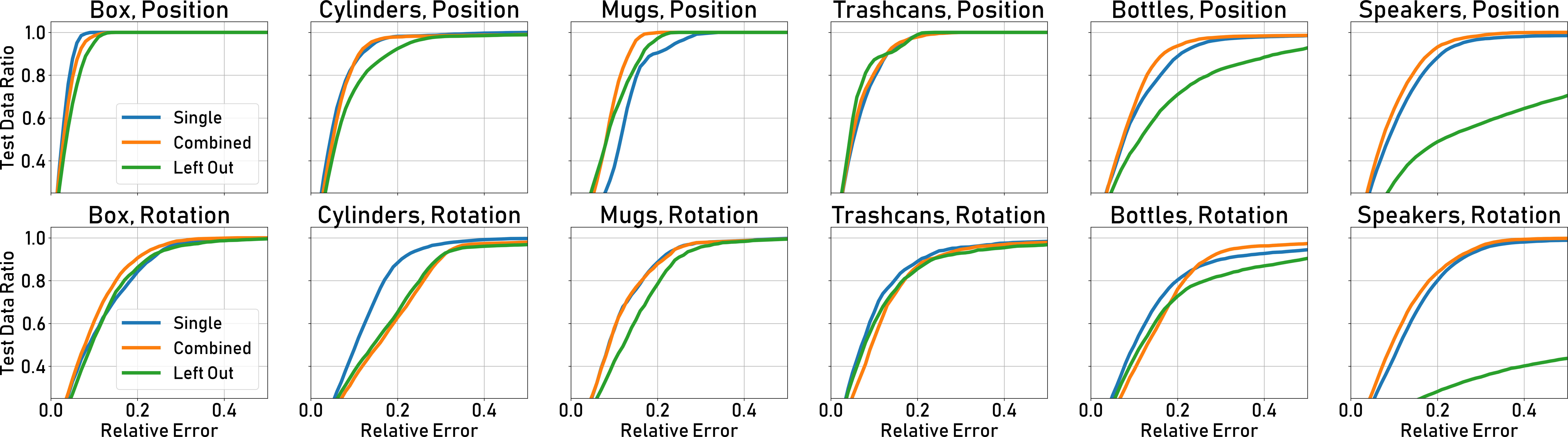}
\end{center}
   \caption{Comparison of performance training on single object categories (blue), the full \texttt{Combined} dataset (orange), and \texttt{Combined} dataset with the evaluated category left out (green). Curves show cumulative fraction of test examples under a certain relative error.}
\label{fig:leaveoutcomp}
\end{figure*}

\subsection{Impulse Generalization}
We first perform impulse generalization (ImpulseGen) experiments to evaluate our model's robustness to \emph{new impulsive forces} applied to known objects---an important capability that helps generalize to novel settings and has been demonstrated in previous work~\cite{byravan2017se3}. For these experiments, train and test sets contain the same objects but with different simulations. Model performance is tested after training on both single object categories and across multiple categories.

\parahead{Single Category}
A separate model is trained for each distinct object category. The second and third columns of Table~\ref{table:resultssummary} show the mean relative percent errors (absolute errors are in brackets) for position and total rotation of the six trained single-category models. As previously hypothesized, the network has a harder time predicting total rotation than position resulting in higher errors. However, having seen all objects during training, these models make extremely accurate predictions on novel impulse forces.

\parahead{Combined Categories}
A single model is trained on the \texttt{Combined} dataset which contains all object categories. We evaluate this model on both the \texttt{Combined} dataset and individual datasets so that performance can be compared to single-category trained models. As shown in columns four and five of Table~\ref{table:resultssummary}, the combined model performs only slightly worse on the individual category datasets compared to single-category training. This suggests the network has effectively learned how varying the force affects angular impulse, and linear and angular velocities in order to perform well on such a wide range of objects.

\subsection{Object Generalization}
We next perform object generalization (ObjGen) experiments to evaluate whether the learned model is able to apply accurate dynamics predictions to unseen objects---a crucial ability for autonomous systems in unseen environments. Since it is impossible to experience all objects that an agent will interact with, we would like knowledge of similarly-shaped objects to inform reasonable predictions about dynamics in new settings. For these experiments, we split datasets based on unique objects such that \textbf{no test objects are seen during training}. Furthermore, the impulses applied for test objects are disjoint from the training objects similar to the ImpulseGen experiments. Since our network is designed specifically to process object shape and learn relevant physical properties, we expect it to extract general features allowing for accurate predictions even on novel objects. Similar to the ImpulseGen experiment, we evaluate models trained on both single and combined categories.

\parahead{Single Category}
Results for testing data when a separate network is trained for each category are shown in Table~\ref{table:resultssummary} under the \textcolor{blue}{blue heading}. As expected, the error is slightly higher but still within 5\% of the ImpulseGen single-category results in most cases. This indicates that the network is able to generalize to unseen objects within the same shape category. The \textcolor{blue}{blue curves} in Figure~\ref{fig:leaveoutcomp} summarize single-category performance over entire test sets. For position, around 90\% of predictions for all object categories fall under 20\% relative error, while for rotation this number falls closer to 80-85\% especially for the larger and more diverse \texttt{Bottles} and \texttt{Speakers} datasets.

\parahead{Combined Categories}
Performance of the model when trained on the \texttt{Combined} dataset and then evaluated on all individual datasets is shown under the \textcolor{orange}{orange heading} in Table~\ref{table:resultssummary}. In general, performance is very similar to training on individual datasets and even improves errors in some cases, for example position predictions for \texttt{Bottles}, \texttt{Mugs}, and \texttt{Speakers}. This indicates that exposing the network to larger shape diversity at training time can help focus learning on underlying physical relationships rather than properties of single or small groups of objects. Improvements and drops in performance are indicated by the blue and \textcolor{orange}{orange curves} plotted in Figure~\ref{fig:leaveoutcomp}. In order to maintain this high performance, the network is likely learning a general approach to extract salient physical features from the diverse objects in the \texttt{Combined} dataset rather than just memorizing how specific shapes behave.

\parahead{Physical \emph{Understanding}}
To explore the implicit feature space of our learned model, we conduct two experiments that require additional supervised outputs from the network on top of position and total rotation (modifications detailed in dashed boxes in Figure~\ref{fig:arch}). In the first experiment, we output the mass and moment of inertia of the object using a feature from the shape processing branch. After training on the combined dataset, the network achieves 8.7\% and 3.5\% relative errors for \textbf{moment of inertia} and \textbf{mass}, respectively, without degrading performance on the main objective. In the second experiment, we supervise \textbf{initial linear} and \textbf{angular velocity} from a feature in the final prediction branch and are able to achieve 3.5\% and 5.8\% relative error, respectively, without affecting the main task. Since there is no significant change in network performance on final state prediction, we conclude that the model has already developed implicit notions of these physical properties, resulting in minimal change to its learned feature space when additional supervised objectives are added.
\begin{figure*}[t]
\begin{center}
   \includegraphics[width=0.98\textwidth]{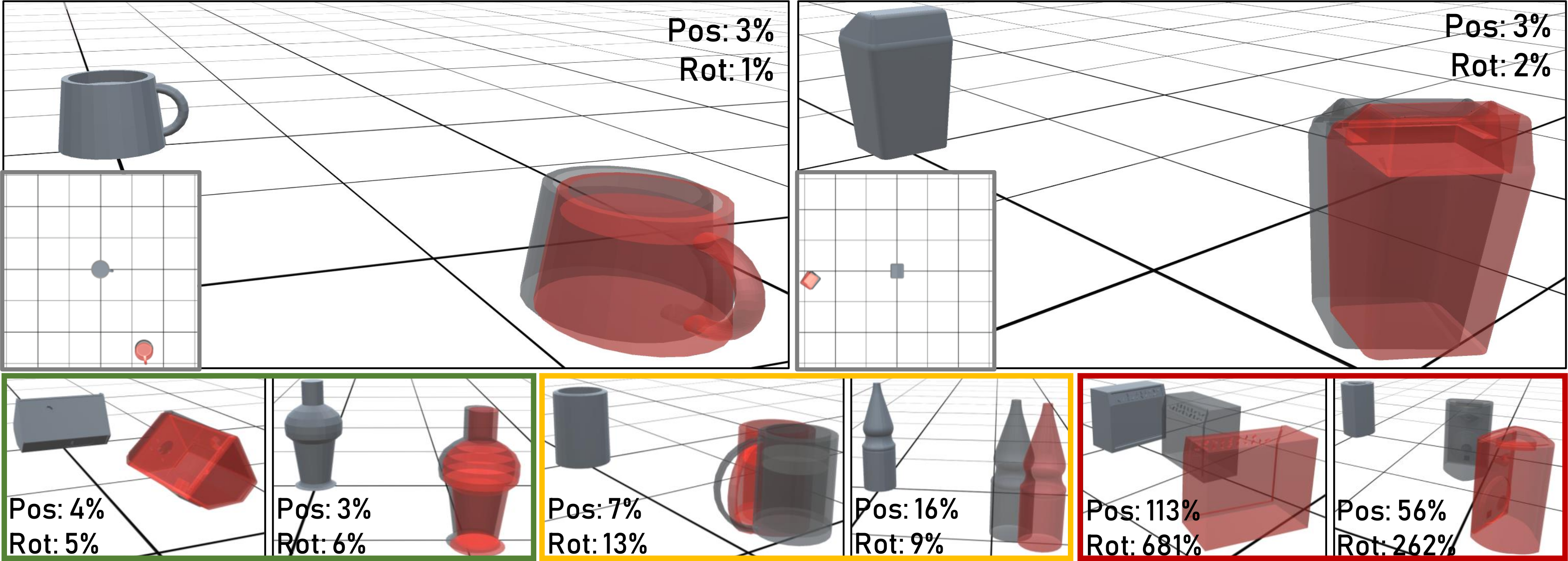}
\end{center}
   \caption{Sample predictions from models trained on the \texttt{Combined} dataset with one category left out. Initial object state is shown in shaded grey, ground truth final state is in transparent grey, and network prediction is in transparent red. Relative errors are shown. From left to right the second row shows best performance (green), average performance (orange), and \texttt{Speakers} failure cases (red).}
\label{fig:predresults}
\end{figure*}

\parahead{Out of Category}
Lastly, we evaluate performance on the extreme task of generalizing \textbf{outside of trained object categories}. For this, we create new combined datasets each with one object category left out of the training set. We then evaluate its performance on objects from the left out category. Results for these experiments are shown under the \textcolor{ForestGreen}{green heading} in Table~\ref{table:resultssummary} and by the \textcolor{ForestGreen}{green curves} in Figure~\ref{fig:leaveoutcomp}. The network is able to achieve good results on all left-out object categories except for \texttt{Speakers}. As seen in Figure~\ref{fig:datadist}, \texttt{Speakers} contributes the most unique objects to the \texttt{Combined} dataset by far; without them, the network may not see enough diversity in training to perform well. Overall, this result shows that we can still make accurate predictions for objects from completely different categories in spite of their shape not being close to the trained objects. The model seems to have developed a deep understanding of how shape affects dynamics through mass, moment of inertia, and contact surface in order to generalize to novel categories at test time. Some predictions from leave-one-out trained models are visualized in Figure~\ref{fig:predresults}.

\begin{figure}[t]
\begin{center}
   \includegraphics[width=\linewidth]{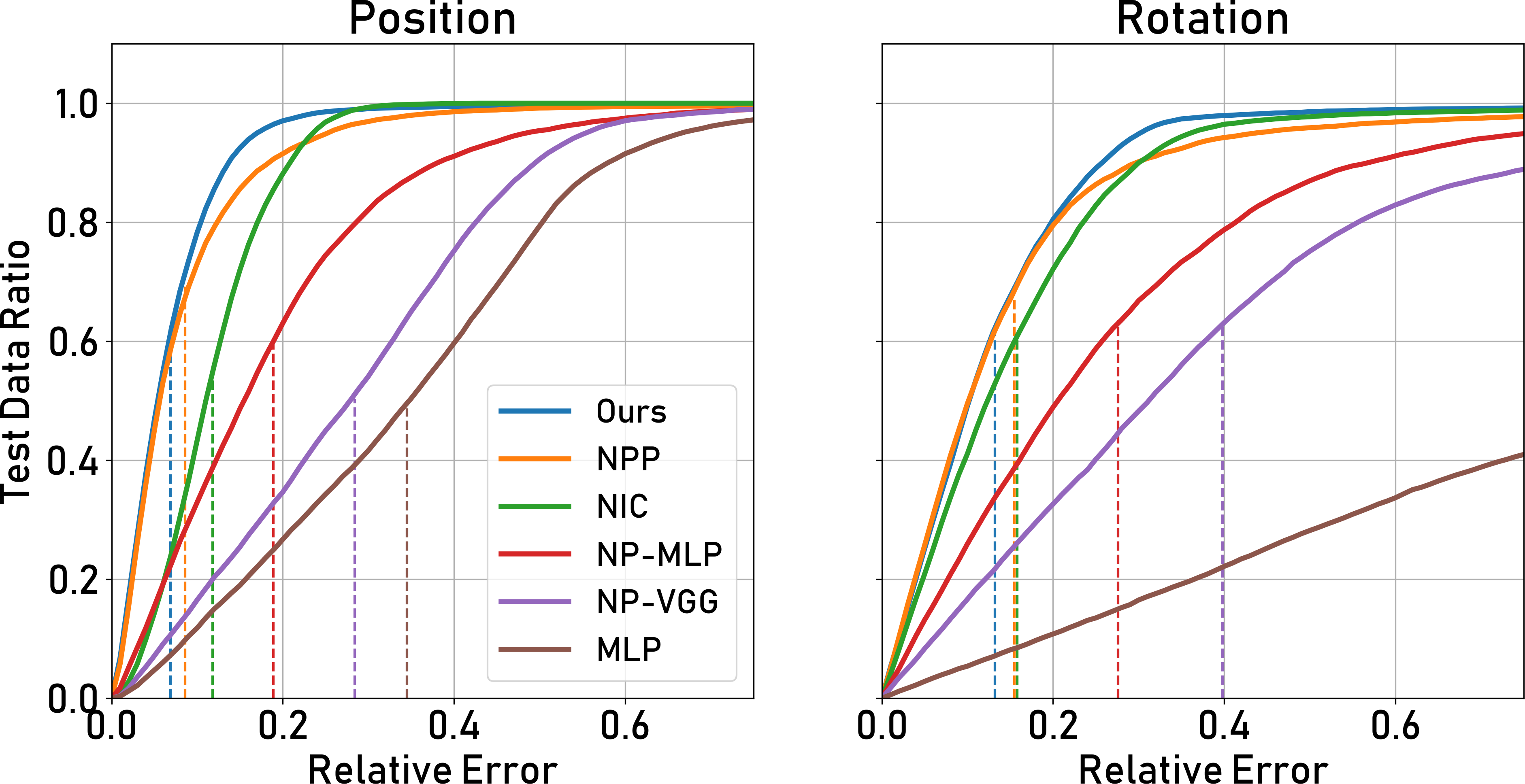}
\end{center}
   \caption{Comparison of architecture variants trained and evaluated on the object generalization \texttt{Combined} dataset. Curves show cumulative fraction of test examples under a certain relative error. Dashed lines indicate the mean relative error. The proposed architecture (Ours) is shown in blue.}
\label{fig:archcomp}
\end{figure}

\subsection{Ablation Study}
We compare our proposed architecture to a number of ablations and modifications to justify design decisions. A performance comparison of all model variants is shown in Figure~\ref{fig:archcomp}. Every model is trained and evaluated on the \texttt{Combined} dataset split by unique objects. The first two ablations justify some physically-informed design choices. The no pairwise products (NPP) model removes the input augmentation of pairwise terms from the impulse processing branch, while the no impulse coordinates (NIC) model uses data that does not use impulse coordinates. The next two models justify our use of PointNet for feature extraction and 3D point clouds as input. In the first model, we replace the PointNet module in the shape processing branch with a feed-forward network (NP-MLP). The second (NP-VGG) shows the advantage of using 3D data to provide shape information instead of images by replacing PointNet with the convolutional layers of VGG-16~\cite{simonyan2014vgg}. This version of the network is trained with images of the objects in each simulation, and performs significantly worse than using point cloud data. This indicates that learning to understand subtle shape variation is imperative to making accurate dynamics predictions for 3D objects. The last model is a straightforward baseline MLP that takes in all input data concatenated together to make position and rotation predictions. The poor performance of this network highlights the advantage of using a branched structure.

\subsection{Comparison to Other Work}
We compare our method to the \emph{hierarchical relation network} (HRN)~\cite{mrowca2018flexible} to highlight the differences between final rest state (our work) and their multi-step predictions. Both models are trained on a small dataset of 1519 scaled boxes simulated in the NVIDIA FleX engine~\cite{macklin2014unified}, then evaluated on 160 held out simulations. Since HRN makes predictions for the next time step, to infer final rest state the model must roll out over roughly 20 steps for each simulation. Our model averages \textbf{6.8\%} and \textbf{11.4\%} relative error for position and rotation, respectively, while HRN achieves \textbf{12.2\%} and \textbf{58.0\%}. It is clear that directly predicting final state allows for more accurate long-term observations, especially for complex rotation motion. However, HRN predicts a detailed trajectory of object motion at 10~Hz which we cannot.

\section{Limitations and Future Work}
\label{sec:limitations}
Our approach has many limitations and there remains room for future exploration. In this work, we took a different approach to previous work by predicting the final state of a 3D rigid object 
instead of multi-step predictions. We believe that future work should consider closing the loop by predicting both the final state as well as multiple intermediate states. Our method is fully supervised and does not explicitly model physical laws like some previous work~\cite{stewart2017label}---we plan to explore this in future work. We show our results on purely synthetic data with no noise and assume that a complete point cloud of an object is available which may not be the case with real-world depth sensing. We ignore the physical parameter estimation problem and assume constant friction and density. We also ignore free 3D dynamics and complex phenomena such as collisions which are important directions for future work. We believe that our approach provides a strong foundation for developing methods for these complex motions.
\section{Conclusion}
We presented a method for learning to predict the position and total rotation of a 3D rigid object subjected to an impulse and moving along a plane. Our method is capable of generalizing to previously unseen object shapes and new impulses not seen during training. We showed that this challenging dynamics prediction problem can be solved using a neural network architecture inspired by physical laws and priors. We train our network on 3D point clouds of a large shape collection and a large synthetic dataset with experiments showing that we are able to accurately predict the final state of 3D rigid objects with complex dynamics.

\parahead{Acknowledgements} 
This work was supported by a grant from the Toyota-Stanford Center for AI Research, NSF grant IIS-1763268, and a Vannevar Bush Faculty Fellowship.
{\small
\bibliographystyle{ieee}
\bibliography{references}

\begin{thebibliography}{10}\itemsep=-1pt

\bibitem{bullet}
Bullet physics engine.
\newblock \url{https://pybullet.org}.

\bibitem{unity}
Unity game engine.
\newblock \url{https://unity3d.com}.

\bibitem{agrawal2016poke}
P.~Agrawal, A.~Nair, P.~Abbeel, J.~Malik, and S.~Levine.
\newblock Learning to poke by poking: Experiential learning of intuitive
  physics.
\newblock In {\em Proceedings of the 30th Conference on Neural Information
  Processing Systems (NIPS)}, 2016.

\bibitem{baillargeon1990top}
R.~Baillargeon and S.~Hanko-Summers.
\newblock Is the top object adequately supported by the bottom object? young
  infants' understanding of support relations.
\newblock {\em Cognitive Development}, 5(1):29--53, 1990.

\bibitem{battaglia2016interactionnets}
P.~Battaglia, R.~Pascanu, M.~Lai, D.~J. Rezende, and K.~kavukcuoglu.
\newblock Interaction networks for learning about objects, relations and
  physics.
\newblock In {\em Proceedings of the 30th International Conference on Neural
  Information Processing Systems (NIPS)}, pages 4509--4517, 2016.

\bibitem{battaglia2013simunderstanding}
P.~W. Battaglia, J.~B. Hamrick, and J.~B. Tenenbaum.
\newblock Simulation as an engine of physical scene understanding.
\newblock {\em Proceedings of the National Academy of Sciences},
  110(45):18327--18332, 2013.

\bibitem{byravan2017se3}
A.~Byravan and D.~Fox.
\newblock Se3-nets: Learning rigid body motion using deep neural networks.
\newblock In {\em 2017 IEEE International Conference on Robotics and Automation
  (ICRA)}, 2017.

\bibitem{chang2015shapenet}
A.~X. Chang, T.~Funkhouser, L.~Guibas, P.~Hanrahan, Q.~Huang, Z.~Li,
  S.~Savarese, M.~Savva, S.~Song, H.~Su, et~al.
\newblock Shapenet: An information-rich 3d model repository.
\newblock {\em arXiv preprint arXiv:1512.03012}, 2015.

\bibitem{chang2017compositional}
M.~B. Chang, T.~Ullman, A.~Torralba, and J.~B. Tenenbaum.
\newblock A compositional object-based approach to learning physical dynamics.
\newblock In {\em Proceedings of the 5th International Conference on Learning
  Representations (ICLR)}, 2017.

\bibitem{ehrhardt2017longterm}
S.~Ehrhardt, A.~Monszpart, N.~{J. Mitra}, and A.~Vedaldi.
\newblock {Learning A Physical Long-term Predictor}.
\newblock {\em arXiv preprint, arXiv:1703.00247}, Mar. 2017.

\bibitem{ehrhardt2018visualobs}
S.~Ehrhardt, A.~Monszpart, N.~J. Mitra, and A.~Vedaldi.
\newblock Unsupervised intuitive physics from visual observations.
\newblock {\em arXiv preprint, arXiv:1805.05086}, 2018.

\bibitem{ehrhardt2017mechanics}
S.~Ehrhardt, A.~Monszpart, A.~Vedaldi, and N.~{J. Mitra}.
\newblock {Learning to Represent Mechanics via Long-term Extrapolation and
  Interpolation}.
\newblock {\em arXiv preprint arXiv:1706.02179}, June 2017.

\bibitem{finn2016videoprediction}
C.~Finn, I.~Goodfellow, and S.~Levine.
\newblock Unsupervised learning for physical interaction through video
  prediction.
\newblock In {\em Proceedings of the 30th International Conference on Neural
  Information Processing Systems (NIPS)}, pages 64--72, 2016.

\bibitem{finn2017planning}
C.~Finn and S.~Levine.
\newblock Deep visual foresight for planning robot motion.
\newblock In {\em International Conference on Robotics and Automation (ICRA)},
  2017.

\bibitem{fraccaro2017disentangled}
M.~Fraccaro, S.~Kamronn, U.~Paquet, and O.~Winther.
\newblock A disentangled recognition and nonlinear dynamics model for
  unsupervised learning.
\newblock In {\em Advances in Neural Information Processing Systems (NIPS)},
  2017.

\bibitem{fragkiadaki2016visualbilliards}
K.~Fragkiadaki, P.~Agrawal, S.~Levine, and J.~Malik.
\newblock Learning visual predictive models of physics for playing billiards.
\newblock In {\em Proceedings of the 4th International Conference on Learning
  Representations (ICLR)}, 2016.

\bibitem{hamrick2016decision}
J.~B. Hamrick, R.~Pascanu, O.~Vinyals, A.~Ballard, N.~Heess, and P.~Battaglia.
\newblock Imagination-based decision making with physical models in deep neural
  networks.
\newblock In {\em Advances in Neural Information Processing Systems (NIPS),
  Intuitive Physics Workshop}, 2016.

\bibitem{kingma2015adam}
D.~P. Kingma and J.~Ba.
\newblock Adam: {A} method for stochastic optimization.
\newblock In {\em International Conference for Learning Representations
  (ICLR)}, 2015.

\bibitem{lerer2016fbtowers}
A.~Lerer, S.~Gross, and R.~Fergus.
\newblock Learning physical intuition of block towers by example.
\newblock In {\em Proceedings of the 33rd International Conference on
  International Conference on Machine Learning (ICML)}, pages 430--438, 2016.

\bibitem{leslie1982perception}
A.~M. Leslie.
\newblock The perception of causality in infants.
\newblock {\em Perception}, 11(2):173--186, 1982.

\bibitem{li2016fall}
W.~Li, S.~Azimi, A.~Leonardis, and M.~Fritz.
\newblock To fall or not to fall: {A} visual approach to physical stability
  prediction.
\newblock {\em arXiv preprint, arXiv:1604.00066}, 2016.

\bibitem{li2017stability}
W.~Li, A.~Leonardis, and M.~Fritz.
\newblock Visual stability prediction for robotic manipulation.
\newblock In {\em 2017 IEEE International Conference on Robotics and Automation
  (ICRA)}, pages 2606--2613, May 2017.

\bibitem{liu2018ppd}
Z.~Liu, W.~T. Freeman, J.~B. Tenenbaum, and J.~Wu.
\newblock Physical primitive decomposition.
\newblock In {\em Proceedings of the 15th European Conference on Computer
  Vision (ECCV)}, 2018.

\bibitem{macklin2014unified}
M.~Macklin, M.~M{\"u}ller, N.~Chentanez, and T.-Y. Kim.
\newblock Unified particle physics for real-time applications.
\newblock {\em ACM Transactions on Graphics (TOG)}, 33(4):153, 2014.

\bibitem{mirza2016unsupervised}
M.~{Mirza}, A.~{Courville}, and Y.~{Bengio}.
\newblock {Generalizable Features From Unsupervised Learning}.
\newblock {\em arXiv preprint, arXiv:1612.03809}, 2016.

\bibitem{mottaghi2016newton}
R.~Mottaghi, H.~Bagherinezhad, M.~Rastegari, and A.~Farhadi.
\newblock Newtonian image understanding: Unfolding the dynamics of objects in
  static images.
\newblock In {\em Proc. Computer Vision and Pattern Recognition (CVPR)}, 2016.

\bibitem{mottaghi2016if}
R.~Mottaghi, M.~Rastegari, A.~Gupta, and A.~Farhadi.
\newblock ``what happens if..." learning to predict the effect of forces in
  images.
\newblock In {\em Proceedings the 14th European Conference on Computer Vision
  (ECCV)}, 2016.

\bibitem{mrowca2018flexible}
D.~Mrowca, C.~Zhuang, E.~Wang, N.~Haber, L.~Fei-Fei, J.~B. Tenenbaum, and
  D.~L.~K. Yamins.
\newblock Flexible neural representation for physics prediction.
\newblock In {\em Proceedings of the 32nd International Conference on Neural
  Information Processing Systems (NIPS)}, 2018.

\bibitem{oh2015atari}
J.~Oh, X.~Guo, H.~Lee, R.~L. Lewis, and S.~P. Singh.
\newblock Action-conditional video prediction using deep networks in atari
  games.
\newblock In {\em Advances in Neural Information Processing Systems (NIPS)},
  2015.

\bibitem{qi2017pointnet}
C.~R. Qi, H.~Su, K.~Mo, and L.~J. Guibas.
\newblock Pointnet: Deep learning on point sets for 3d classification and
  segmentation.
\newblock {\em Proc. Computer Vision and Pattern Recognition (CVPR), IEEE},
  1(2):4, 2017.

\bibitem{riochet2018intphys}
R.~Riochet, M.~Y. Castro, M.~Bernard, A.~Lerer, R.~Fergus, V.~Izard, and
  E.~Dupoux.
\newblock Intphys: {A} framework and benchmark for visual intuitive physics
  reasoning.
\newblock {\em arXiv preprint, arXiv:1803.07616}, 2018.

\bibitem{imagenet}
O.~Russakovsky, J.~Deng, H.~Su, J.~Krause, S.~Satheesh, S.~Ma, Z.~Huang,
  A.~Karpathy, A.~Khosla, M.~Bernstein, A.~C. Berg, and L.~Fei-Fei.
\newblock {ImageNet Large Scale Visual Recognition Challenge}.
\newblock {\em International Journal of Computer Vision (IJCV)},
  115(3):211--252, 2015.

\bibitem{sanchez2018graphnet}
A.~{Sanchez-Gonzalez}, N.~{Heess}, J.~T. {Springenberg}, J.~{Merel},
  M.~{Riedmiller}, R.~{Hadsell}, and P.~{Battaglia}.
\newblock {Graph networks as learnable physics engines for inference and
  control}.
\newblock In {\em Proceedings the 35th International Conference on Machine
  Learning (ICML)}, 2018.

\bibitem{simonyan2014vgg}
K.~Simonyan and A.~Zisserman.
\newblock Very deep convolutional networks for large-scale image recognition.
\newblock {\em CoRR}, abs/1409.1556, 2014.

\bibitem{stewart2017label}
R.~Stewart and S.~Ermon.
\newblock Label-free supervision of neural networks with physics and domain
  knowledge.
\newblock In {\em Proc. of AAAI Conference on Artificial Intelligence}, 2017.

\bibitem{wang2018physnet}
Z.~Wang, S.~Rosa, B.~Yang, S.~Wang, N.~Trigoni, and A.~Markham.
\newblock 3d-physnet: Learning the intuitive physics of non-rigid object
  deformations.
\newblock In {\em Proceedings of the 26th International Joint Conference on
  Artificial Intelligence, {IJCAI-18}}, pages 4958--4964, 2018.

\bibitem{watters2017vin}
N.~Watters, A.~Tacchetti, T.~Weber, R.~Pascanu, P.~Battaglia, and D.~Zoran.
\newblock Visual interaction networks.
\newblock {\em arXiv preprint, arXiv:1706.01433}, 2017.

\bibitem{wu2016phys101}
J.~Wu, J.~J. Lim, H.~Zhang, J.~B. Tenenbaum, and W.~T. Freeman.
\newblock Physics 101: Learning physical object properties from unlabeled
  videos.
\newblock In {\em Proceedings of the 27th British Machine Vision Conference
  (BMVC)}, 2016.

\bibitem{wu2017deanimation}
J.~Wu, E.~Lu, P.~Kohli, W.~T. Freeman, and J.~B. Tenenbaum.
\newblock Learning to see physics via visual de-animation.
\newblock In {\em Proceedings of the 31st Conference on Neural Information
  Processing Systems (NIPS)}, 2017.

\bibitem{wu2015galileo}
J.~Wu, I.~Yildirim, J.~J. Lim, W.~T. Freeman, and J.~B. Tenenbaum.
\newblock Galileo: Perceiving physical object properties by integrating a
  physics engine with deep learning.
\newblock In {\em Proceedings of the 29th Conference on Neural Information
  Processing Systems (NIPS)}, pages 127--135, 2015.

\bibitem{wu2015modelnet}
Z.~Wu, S.~Song, A.~Khosla, F.~Yu, L.~Zhang, X.~Tang, and J.~Xiao.
\newblock 3d shapenets: A deep representation for volumetric shapes.
\newblock In {\em Computer Vision and Pattern Recognition (CVPR)}, 2015.

\bibitem{ye2018interpretable}
T.~Ye, X.~Wang, J.~Davidson, and A.~Gupta.
\newblock Interpretable intuitive physics model.
\newblock In {\em Proceedings the 15th European Conference on Computer Vision
  (ECCV)}, pages 89--105, 2018.

\bibitem{zhang2016blocks}
R.~Zhang, J.~Wu, C.~Zhang, W.~T. Freeman, and J.~B. Tenenbaum.
\newblock A comparative evaluation of approximate probabilistic simulation and
  deep neural networks as accounts of human physical scene understanding.
\newblock In {\em Annual Meeting of the Cognitive Science Society}, 2016.

\end{thebibliography}
}

\renewcommand\thesection{\Alph{section}}
\setcounter{section}{0}
\renewcommand\thefigure{\thesection.\arabic{figure}}    
\setcounter{figure}{0} 

\section{Appendix: Implementation Details}
\label{sec:implementation}
Here we provide additional details of our simulation pipeline and baseline implementations.

\parahead{Simulation Procedure}
The simulation pipeline is introduced in Section 4 of the main paper. Prior to simulation, some pre-computation is done on object shapes to extract accurate physical parameters, namely the mass and moment of inertia. To calculate these values, we voxelize each shape using a grid with a cell side length of 2.5 cm. From this we approximate the volume of the object which can be used to calculate the mass given density. Additionally, we compute a discretized approximation of the moment of inertia about the shape's principle axes. These calculated mass and moment of inertia values are used directly to parameterize the simulated rigid bodies within the the Bullet physics engine. Calculating physical parameters in this way ensures consistency across all simulated shapes rather than relying on the physics engine to calculate them using a mesh collider which can be extremely inconsistent and inaccurate. 

For each performed simulation, the amplitude of the randomly applied impulse is scaled by the object mass to ensure similar distributions for small and large objects alike. When choosing the impulse to apply, we also ensure that the line defined by the impulse vector and its position passes within a certain radius of the object's center of mass. Constraining the impulse in this way gives control over the amount of rotation the object will undergo, allowing for a reasonable distribution. For \texttt{Speakers} this radius is 40 cm because these objects tend to have a large moment of inertia about the vertical axis. For all other object categories this radius is set to 10 cm. Simulated objects have a friction coefficient of 0.7 in the Bullet physics engine while the ground plane has a coefficient of 0.1. 

For each object, we also sample a point cloud from the surface to use as input to our model. To ensure uniform sampling, we first oversample (by a factor of 3) the mesh surface area. We then sub-sample these points using furthest point sampling to obtain our final 1024 points.

\parahead{Baselines}
We now detail implementations for the baselines presented in Section 6.3 of the main text. For the no PointNet MLP (NP-MLP) baseline, we replace the PointNet module with a small MLP made up of 2 fully-connected (FC) layers each with 512 nodes. This is followed by the usual shape processing branch.

\begin{figure}[t]
\begin{center}
   \includegraphics[width=\linewidth]{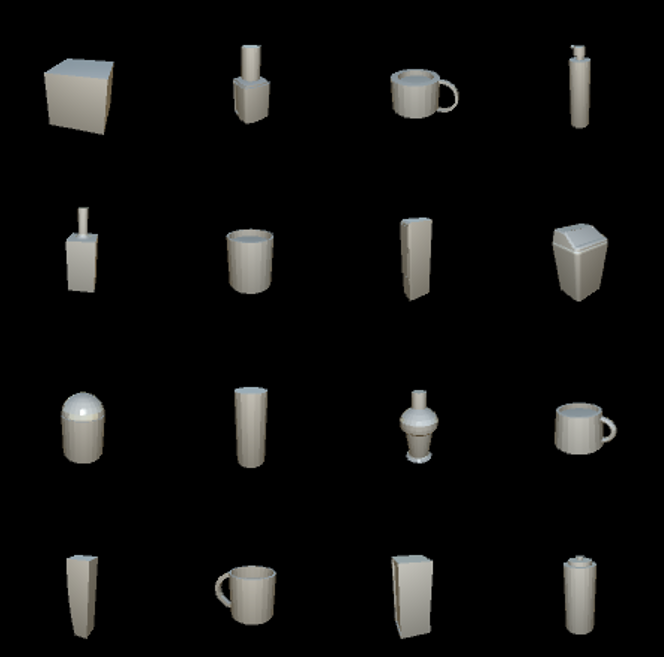}
\end{center}
   \caption{Examples of object images used as input to the NP-VGG baseline model.}
\label{fig:vggimages}
\end{figure}

For the no PointNet VGG (NP-VGG) baseline, we replace the PointNet module with a modified version of VGG-16~\cite{simonyan2014vgg}. We use all convolutional layers from VGG (up through the 5th max pooling layer), but modify the final 2 FC layers to each have 1024 nodes. We initialize the weights of the convolutional layers from a model pretrained on ImageNet~\cite{imagenet}. During training, we fine tune the weights for our problem. The input to NP-VGG is a single 112$\times$112 image of the simulated object rather than a point cloud (see Figure~\ref{fig:vggimages}). In these images, the object is placed against a plain black background and the camera is positioned in line with the applied impulse (offset in the -x direction in the impulse coordinate system described in Section 4 of the main paper).

The basic MLP baseline flattens the point cloud input and concatenates it to all other inputs to process this whole vector with a series of FC layers: 512x3 (3 layers with output size 512), 256x3, 128x2, 64x1, 32x1, and then the final output of 3 (2 position and 1 total rotation). The poor performance of this baseline is likely due to the lack of a branched structure which overwhelms the network with point cloud information making it difficult to pick out the important impulse input.

\setcounter{figure}{0} 

\section{Appendix: Additional Results}
\label{sec:addedresults}
In this section we provide some additional results to those presented in Section 6 of the main paper.

\begin{figure*}[t]
\begin{center}
   \includegraphics[width=0.98\textwidth]{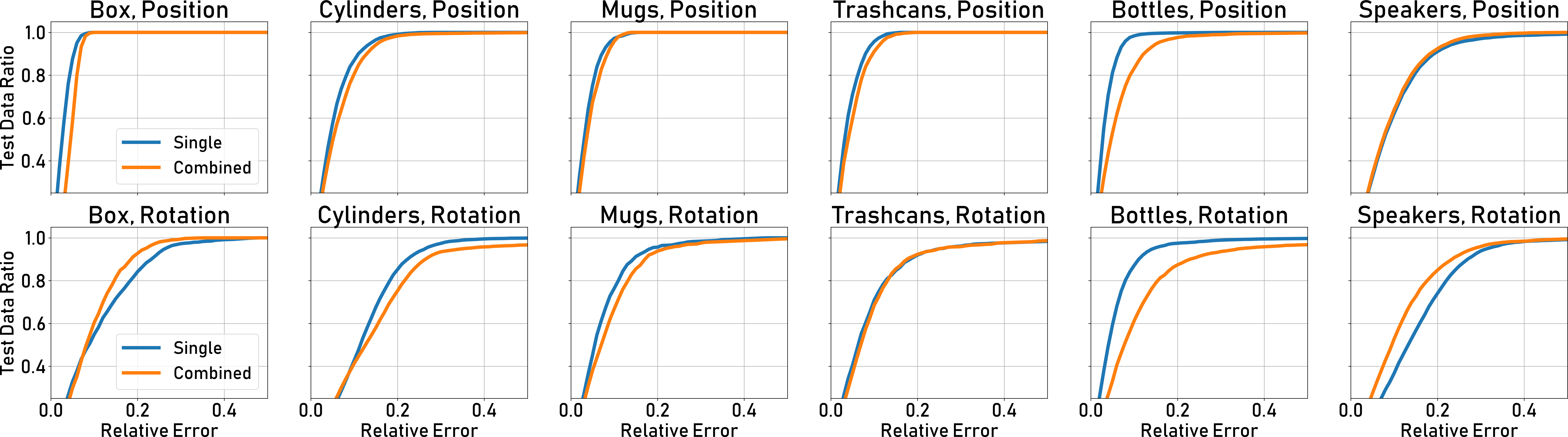}
\end{center}
   \caption{Comparison of performance training on single object categories (blue) and the full \texttt{Combined} dataset (orange) for impulse generalization experiments. Curves show the cumulative fraction of test examples under a certain relative error.}
\label{fig:impulsecurves}
\end{figure*}

\parahead{Impulse Generalization}
Figure~\ref{fig:impulsecurves} summarizes the performance of our method evaluated on novel impulse forces applied to known objects (the ImpulseGen experiments presented in the main paper). These curves plot the fraction of test examples (y-axis) for which model predictions fall under a certain relative error threshold (x-axis). The blue curves indicate models that were trained on datasets of individual objects categories (six different models, one for each column). The orange curves show performance of a single model trained on the \texttt{Combined} dataset then evaluated on each individual category separately. In some cases (\texttt{Box} and \texttt{Speakers} rotation), the combined training offers improved performance, while in others (\texttt{Bottles}) the model seems to benefit from training on individual categories. 

\begin{figure*}[t]
\begin{center}
   \includegraphics[width=0.98\textwidth]{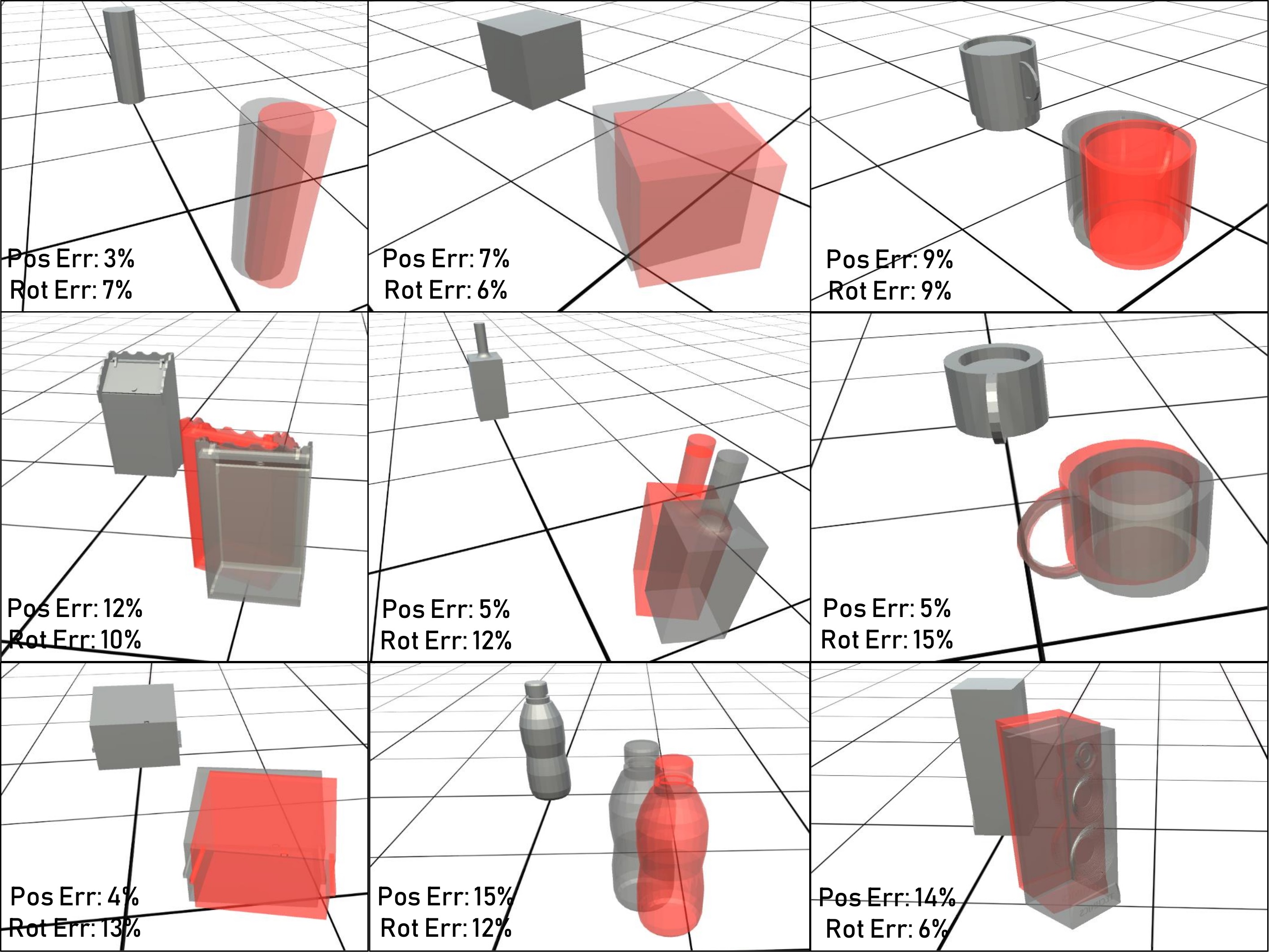}
\end{center}
   \caption{Sample predictions for our model trained on the \texttt{Combined} dataset with one category left out. Initial object state is shown in shaded grey, ground truth final state is in transparent grey, and network prediction is in transparent red. Relative errors are shown.}
\label{fig:predresults}
\end{figure*} 

\begin{figure*}[t]
\begin{center}
   \includegraphics[width=0.9\textwidth]{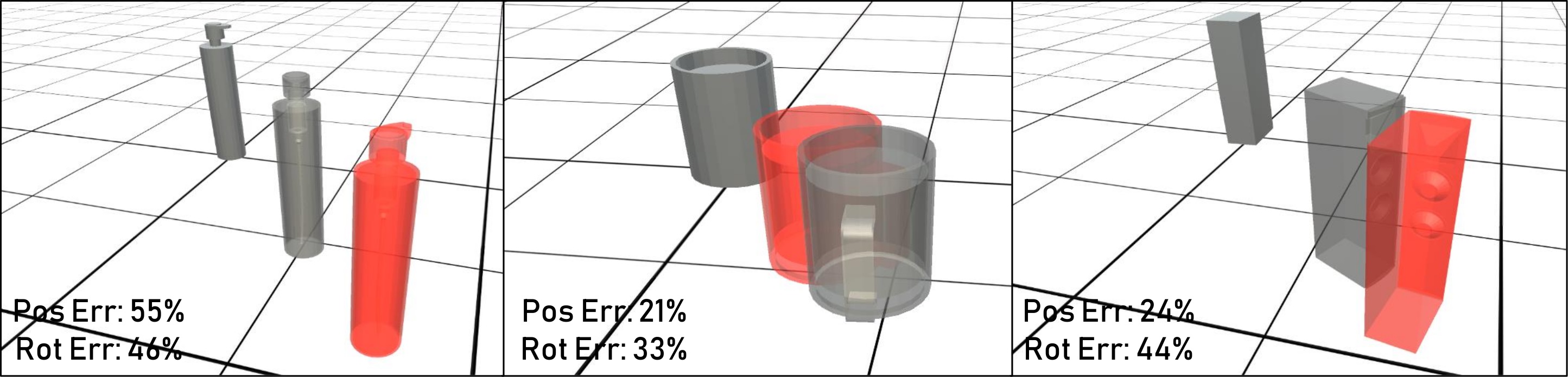}
\end{center}
   \caption{Sample failure cases for our model trained on the \texttt{Combined} dataset with one category left out. Initial object state is shown in shaded grey, ground truth final state is in transparent grey, and network prediction is in transparent red. Relative errors are shown.}
\label{fig:failcases}
\end{figure*} 

\parahead{Out of Category Object Generalization}
Additional visualizations for the out-of-category generalization experiments are shown in Figure~\ref{fig:predresults}. For each of these examples, the model was trained on the \texttt{Combined} dataset with the evaluated object category left out. Some failure cases are presented in Figure~\ref{fig:failcases}.

\begin{figure*}[t]
\begin{center}
   \includegraphics[width=0.9\textwidth]{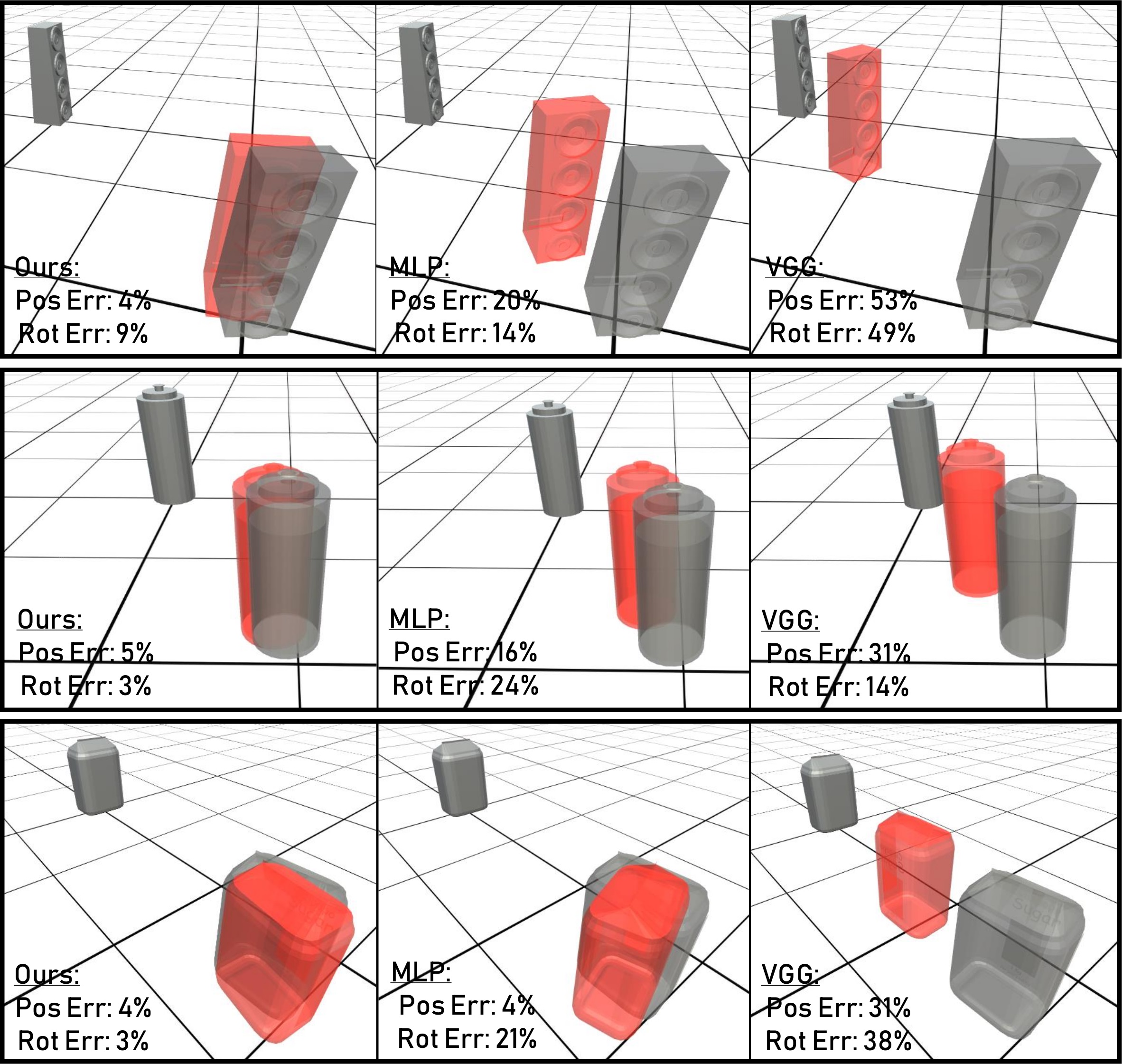}
\end{center}
   \caption{Sample predictions from our model trained on the \texttt{Combined} dataset with one category left out, the NP-MLP baseline trained on the full \texttt{Combined} dataset, and the NP-VGG baseline trained on the full \texttt{Combined} dataset. Initial object state is shown in shaded grey, ground truth final state is in transparent grey, and network prediction is in transparent red. Relative errors are shown. Each row shows a single simulation example.}
\label{fig:baselinecomp}
\end{figure*}

\parahead{Comparison to Baselines}
Figure~\ref{fig:baselinecomp} compares results from our out-of-category trained model to that of two baselines presented in the main text---the no PointNet MLP (NP-MLP) and VGG (NP-VGG) models. Note that our model was trained on the \texttt{Combined} dataset with the evaluated category left out, while the two baseline methods were trained on the full \texttt{Combined} dataset split by objects. Each row shows predictions for each model on the same simulation. We see that our proposed method provides more accurate predictions than both of these baselines even though it did not see the test object category at training time. 

\end{document}